\documentclass[letterpaper, 10 pt, conference]{ieeeconf}  % Comment this line out if you need a4paper

\IEEEoverridecommandlockouts                              % This command is only needed if
\usepackage{graphicx}
\usepackage{amsmath}
\usepackage{amssymb}
\usepackage{breqn}
\usepackage{subcaption}
\usepackage{algorithm}
\usepackage{algorithmicx,algpseudocode}
\algrenewcommand\algorithmicindent{0.6em}
\usepackage{pgfplots}
\usepackage{booktabs, dcolumn}

\pgfplotsset{
every axis/.append style={
  axis line style={->}, % arrows on the axis
  xlabel near ticks,
  ylabel near ticks,
  legend style={font=\scriptsize},
  label style={font=\scriptsize},
  tick label style={font=\scriptsize},
  }
}
\pgfplotsset{compat=newest}
\pgfplotsset{plot coordinates/math parser=false}
\newlength\figureheight
\newlength\figurewidth

\usepackage{stork-hang}

\makeatletter
\let\NAT@parse\undefined
\makeatother
\usepackage[numbers,sort,compress]{natbib}
\usepackage{hyperref}
\usepackage[font=small]{caption}
\usepackage{xcolor,colortbl}
\title{\LARGE \bf%
Rearrangement with Nonprehensile Manipulation Using\\ Deep Reinforcement Learning
}

\author{Weihao Yuan$^{1}$, Johannes A. Stork$^{2}$, Danica Kragic$^{2}$, Michael Y. Wang$^{1}$ and Kaiyu Hang$^{1}$% <-this % stops a space
\thanks{$^1$ These authors are with the Hong Kong University of Science and Technology and HKUST Robotics Institute. W. Yuan is with the Department of Electronic and Computer Engineering. M. Y. Wang is with the Department of Mechanical and Aerospace Engineering and the Department of Electronic and Computer Engineering. K. Hang is with the Department of Computer Science and Engineering and HKUST Institute for Advanced Study.}%
\thanks{$^2$ J. A. Stork and D. Kragic are with the Robotics, Perception and Learning Lab, Centre for Autonomous Systems, KTH Royal Institute of Technology, Sweden}%
}

\begin{document}

\maketitle
% % \thispagestyle{empty}
% % \pagestyle{empty}

%%%%%%%%%%%%%%%%%%%%%%%%%%%%%%%%%%%%%%%%%%%%%%%%%%%%%%%%%%%%%%%%%%%%%%%%%%%%%%%%

%%%%%%%%%%%%%%%%%%%%%%%%%%%%%%%%%%%%%%%%%%%%%%%%%%%%%%%%%%%%%%%%%%%%%%%%%%%%%%%%
% !TEX root =  ../main.tex
\begin{abstract}
%\direction{Write one statement about the general problem}
Rearranging objects on a tabletop surface by means of nonprehensile manipulation is a task which requires skillful interaction with the physical world. Usually, this is achieved by precisely modeling physical properties of the objects, robot, and the environment for explicit planning. 
%\direction{Tell what this paper is about}
In contrast, as explicitly modeling the physical environment is not always feasible and involves various uncertainties, we learn a nonprehensile rearrangement strategy with deep reinforcement learning based on only visual feedback.
%\direction{Describe how it solves the problem}
For this, we model the task with rewards and train a deep $Q$-network. Our potential field-based heuristic exploration strategy reduces the amount of collisions which lead to suboptimal outcomes and we actively balance the training set to avoid bias towards poor examples.
%\direction{Emphasize what is new or better}
Our training process leads to quicker learning and better performance on the task as compared to uniform exploration and standard experience replay.
%\direction{Mention the evidence indicating the advantages of the proposed approach}
We demonstrate empirical evidence from simulation that our method leads to a success rate of $85\%$, show that our system can cope with sudden changes of the environment, and compare our performance with human level performance.
%\todo{Re-write this into something claiming less. Possibly argue via the task: e.g. learning vs. planning etc.}
\end{abstract}
% !TEX root =  ../main.tex

% \todo{There are so many $\alpha$s and other variables used over and over again. }

\section{INTRODUCTION}
% \direction{Start with a motivation}
% -- A Birdview
The skill of \emph{rearrangement planning} is essential for robots for manipulating objects in cluttered and unstructured environments \cite{cosgun11pushplanning, dogar12clutter, gupta12bricksorting, stilman04namo}. Classic approaches to object rearrangement use so-called \emph{pick-and-place actions} and rely on grasp \cite{hang2017framework, hang2016hierarchical, hang2014combinatorial, hang2014hierarchical} and motion planning \cite{simeon04prm, stilman07namo}. Assuming that the robot's workspace is constrained to a tabletop, more recent works try to leverage on \emph{nonprehensile actions} for more efficient solutions \cite{haustein2015kinodynamic, king2015nonprehensile, king2016, king2017}, however, exchanging complex grasp planning for planning of complex robot-object interactions.

% -- The challenges
Besides the fact that the general problem is \emph{NP-hard} \cite{wilfong91planning},  rearrangement planning poses many other challenges which are often addressed under simplified assumptions. Due to occlusions caused by clutter in a single camera setup, a robot often suffers from incomplete knowledge of the environment's state \cite{Schiebener13}. Therefore, a number of recent works assume complete observability of the state from perfect visual perception for planning \cite{haustein2015kinodynamic, king2015nonprehensile, king2016, king2017}. Often, the complex dynamics of nonprehensile interaction are reduced to a quasi-static model \cite{zhou17, fazeli2015} which conveniently allows solutions based on motion primitives \cite{cosgun11, dogar11pushgrasping}. Moreover, for keeping planning of action sequences tractable, physical properties are often assumed to be known such that robot-object interactions can be simulated \cite{king2015nonprehensile}. In some cases, a free-floating end-effector is assumed to avoid expensive planning in configuration space and to allow physics-aware planning with kinodynamic-RRT \cite{haustein2015kinodynamic}. All these approaches treat perception, action planning, and motion planning separately.

\begin{figure}[t]
\centering
\includegraphics[width=0.8\columnwidth, trim={0.2cm 3.8cm 4cm 0cm}, clip]{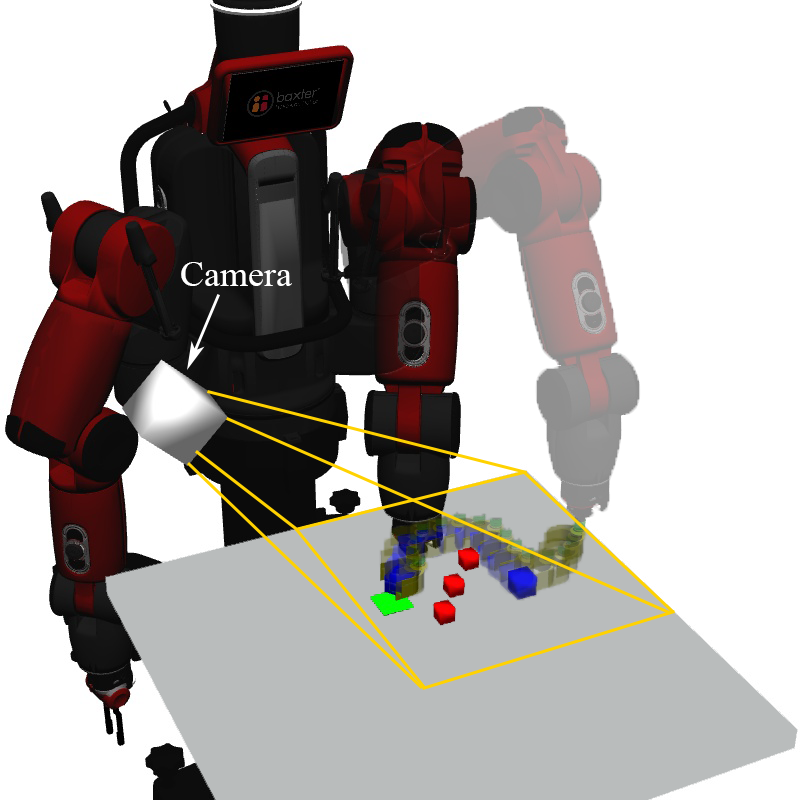}
\caption{The robot is tasked to first find and then push an object (blue) around obstacles (red) to a goal region (green) relying on only visual feedback.}
\label{fig:first}
\vspace{-0.6cm}
\end{figure}

% -- Problem Statement
In this work, we design a learning system that treats perception, action planning, and motion planning in an end-to-end process. Different from whole-arm interaction as studied by King \emph{et al.} \cite{king2015nonprehensile}, our task consists of pushing a manipulation object to a target position while avoiding collisions, as illustrated in Fig. \ref{fig:first}. Perceptions are single-view RGB images and the actions move a manipulation tool in five different directions. Different to model or simulation-based approaches \cite{king2015nonprehensile, haustein2015kinodynamic}, we assume no prior knowledge of any physical properties such as  mass, inertia matrices, friction coefficients, and so on.

% -- Why DQN
Instead of a classic planning framework which requires an explicit physical model, we use model-free $Q$-learning \cite{Sutton98} to find an \emph{optimal} policy directly from visual input. Since our workspace consists of many objects located at arbitrary positions, the state space is infeasible for classic $Q$-learning. However, based on only visual input, latest research on deep $Q$-network (DQN) successfully shows the power of deep convolutional neural networks in playing Atari games with human-level performance \cite{mnih2015human}. Therefore, we employ DQN for our tabletop rearrangement problem, which bears similarities to Atari games, both in perception and state transitions. Similar to the games, our robot operates in a stochastic world where obstacles can move at any time and friction varies, requiring reactive behavior. This can be addressed since a DQN determines actions based on only the current input as opposed to a classic planning framework.

%\subsubsection*{Our contributions}
\textbf{Our contributions} concern both the rearrangement task and the learning process and consist of:
\begin{enumerate}
\item
modeling the rearrangement task as a reinforcement learning problem with task-specific reward functions,
\item
improving the training process by active control of the replay dataset to avoid bias towards suboptimal examples,
\item
devising an informed exploration process based on a Gaussian potential field to reduce the amount of suboptimal examples caused by collisions.
\end{enumerate}
%
%DQNs are trained with \emph{experience reply} from representative replay buffer to counter instability and divergence of the $Q$-function \cite{mnih2015human, mcclelland1995there, o2010play, lin1993reinforcement}. In our task, however, collisions with obstacles lead a biast training set with suboptimal examples. We introduce stochastically control of the reply buffer to balance between success and failure experiences. Since random exploration causes collisions which bias the replay buffer, we propose a Gaussian potential field-based method for informed action sampling \cite{Zhu06, Varava-RSS-17}, in order to further facilitate an efficient buffer control.

In our simulation-based evaluation, the DQN trained with only $2$ random obstacles can achieve high success rates when presented with $2$ to $4$ randomly positioned obstacles. We interpret this as evidence that the network learns both global features for path planning and local features for collision avoidance. Our comparison against the performance of a human expert player indicates that the DQN plans more conservative to avoid collisions. Furthermore, we qualitatively show that our system can react to sudden changes in the positions of the object, obstacles or the target, as well as the randomly altered friction coefficient, and a distracting novel object introduced to the scene.

This paper is structured by formally defining the problem in Sec.~\ref{sec:problem}, and then introducing the necessary preliminaries in Sec.~\ref{sec:background}. In Sec.~\ref{sec:method} we explain the details of our design of DQN-based learning architecture. Finally, we evaluate our system in Sec.~\ref{sec:experiments} and conclude in Sec.~\ref{sec:conclusion}.

% \subsection{Motivation}
% Power of deep learning
%
% Applications of rearrangement manipulation
%
% \direction{Tell what this paper is about}
%
% \direction{Explain what makes this work relevant}

% \subsection{Contribution}
%
% Most works about the rearrangement manipulation before are using complex planning methods, which need much information about the environment and are time-consuming. Different from that, our work is an end-to-end process. With one picture taken from the RGB camera, robot can output its next action. This is a real-time planning and can handle many emergency situations and complex environment beyond the priori knowledge.
%
% \todo{The section above sounds like the claim that planning can be replaced by this approach. Is that the intention? Planning corresponds to learning the policy, the policy corresponds to the plan (formally a conditional plan). Technically, executing the policy is not real-time planning.}
%
% To realize the self-learning of robot, we introduce the reinforcement learning. To make the training in large state space possible, we introduce replay buffer control, piecewise network updating and guided exploration to generate better training data and make better use of them.

% \direction{Structure of this paper (optional)}

% \input{includes/related_work}
% !TEX root =  ../main.tex

\section{PROBLEM STATEMENT}
\label{sec:problem}

% We consider the problem of rearranging an object on a tabletop work-surface to a target location. Provided with only visual feedback, we move the object using a non-prehensile manipulation tool while the obstacles have to be avoided. For this, we learn a function from experience that takes raw video images as input and outputs the utility of each possible tool action.

In this section, we formally define the task and the necessary assumptions.

% \subsection{Assumptions}
\subsection{Task and Assumptions}
\label{sec:problem-assumptions}

We assume that a robot is equipped with a non-prehensile manipulation tool, which can move along the planar work-surface to reach all required positions. As shown in Fig.~\ref{fig:first}, on the work-surface there is a cube-shaped manipulation object, a few cube-shaped obstacle objects, and a squared visual indicator for the target location. The manipulation tool has a fixed orientation, while the target location and the manipulation object on the work-surface are initially situated in the half-space in front of the tool. Mass and friction of the object and obstacles are not known but allow for effortless manipulation. We assume that the target position is not fully blocked by obstacles and that there exists at least one path allowing the manipulation tool to push the object into the target area.

The work-surface is observed by a static single-view RGB camera perceiving the manipulation object in \emph{blue}, the obstacles in \emph{red}, the target location in \emph{green}, and the robot arm with the attached manipulation tool with possible occlusions. Manipulation is done in discrete time steps such that a camera image is recorded at time step $t$ and then an action is executed leading to the next time step $t+1$.

\begin{figure}[]
\centering
\includegraphics[width=0.9\columnwidth]{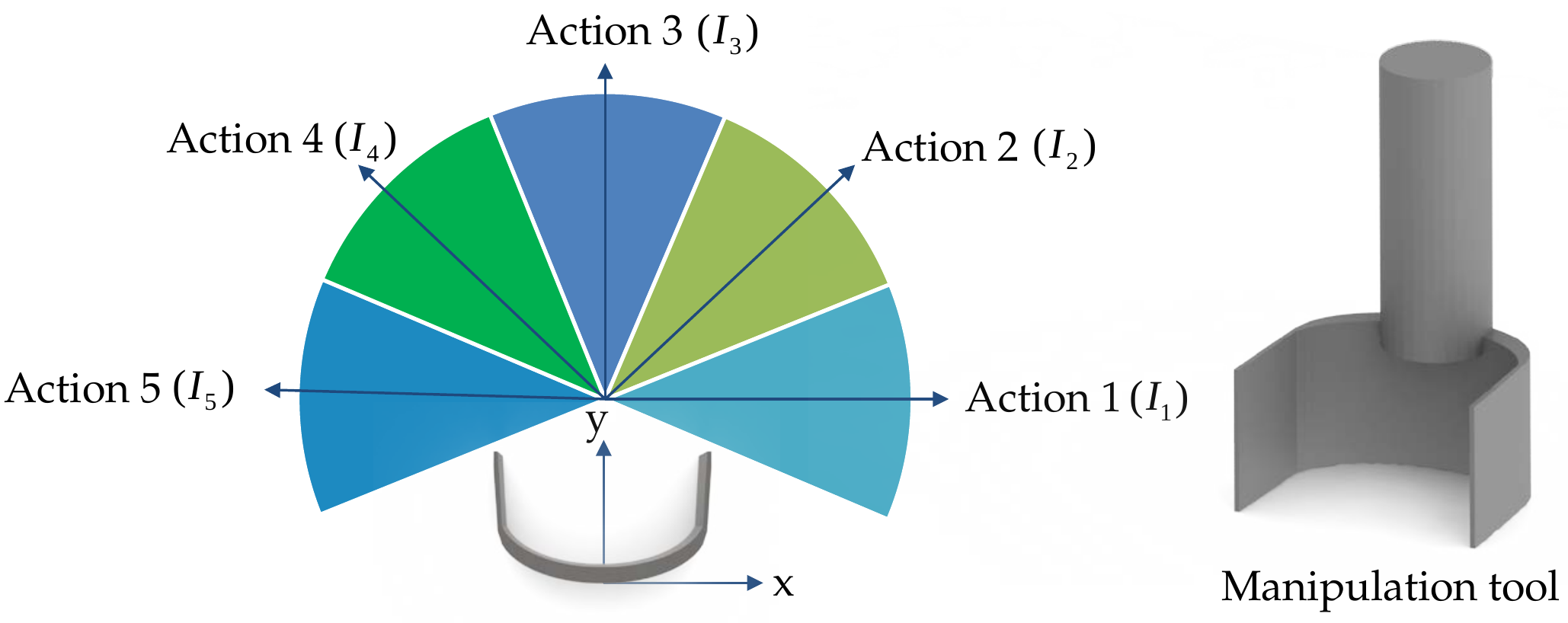}
\caption{This figure illustrates the $5$ predefined motion directions for the manipulation tool. Action $3$ is aligned with the front direction $Y$ of the manipulation tool. The colored sectors depict the ranges in which the potential integrals are calculated, as required in Sec.~\ref{sec:exploration} for informed action sampling.}
\label{fig:actions}
\vspace{-0.6cm}
\end{figure}

The task is to find a sequence of predefined actions, as depicted in Fig.~\ref{fig:actions}, to push the object from a random initial position to the target area while avoiding collisions with any randomly positioned obstacles. Note that with these $5$ actions, the manipulation tool can achieve a large set of trajectories but it cannot move backwards.

% \todo{I think we have to assume that a path exists, that we assume the problem can be solved within a certain number of steps making the problem episodic.}
%
% \todo{Some positions are fixed.}

\subsection{Definitions and Notations}
\label{sec:problem-definitions}

\textbf{Observations.} An observation $\mathbf{x}$ is a $128 \times 128$ RGB image ($49152$ dimensions) taken from the camera pointing at the work-surface. An example of an observation is seen in Fig.~\ref{fig:network_structure}. In the images, the robot and objects can occlude parts of the scene.

\textbf{Actions.} An action $a \in \mathcal{A}$ translates the manipulation tool parallel to the work-surface using one of the predefined motion directions for a fixed step size $d_a\in\mathbb{R}^+$.

\textbf{Episodes.} An episode $E$ is a sequence of actions that is terminated by either success or failure. We index the set of episodes by $k$ and use time steps $t = 1, 2, \dots, T$ within episodes where $T$ might be different from episode to episode.

\textbf{Success and Failure.} An episode terminates with \emph{success} iff. the manipulation object reaches the target location. Otherwise, it terminates with \emph{failure} in cases when too many time steps have passed, obstacles are moved (collision), or the tool is moved outside of the work-surface.

\textbf{Grounding Labels.} During the learning process, the algorithm has access to the following 2D positions relative to the work-surface's frame: Manipulation object position $\mathbf{p}^{\text{man}}$, tool position $\mathbf{p}^{\textrm{tool}}$, target location $\mathbf{p}^{\textrm{target}}$, and for each obstacle $i$ the position $\mathbf{p}^{\textrm{obs}, i}$. The positions are all measured in centimeters. From these we can derive predicates for \emph{success} and \emph{failure}.

% \todo{The reason we train in simulation is that these labels would be difficult to get in real. Mention this somewhere.}

\subsection{Objective}
\label{sec:problem-objective}

Our goal is to learn a robust function $Q(\mathbf{x}, a)$ over all relevant camera images $\mathbf{x}$ and actions $a \in \mathcal{A}$, such that repeatedly taking the best actions $a^* = \arg \max_{a \in \mathcal{A}} Q(\mathbf{x}, a)$ in subsequent situations moves the manipulation object to the target location. It must be possible to start in any situation where the manipulation object and target location are situated in front of the manipulator as described above. Learning this function alleviates the problems of explicitly modeling the environment with its dynamics, tracking the manipulation object, or executing a planning algorithm.

% \todo{Include number of steps requirement here.}

\section{PRELIMINARIES}
\label{sec:background}

\begin{figure}[]
\centering
\includegraphics[width=0.99\columnwidth]{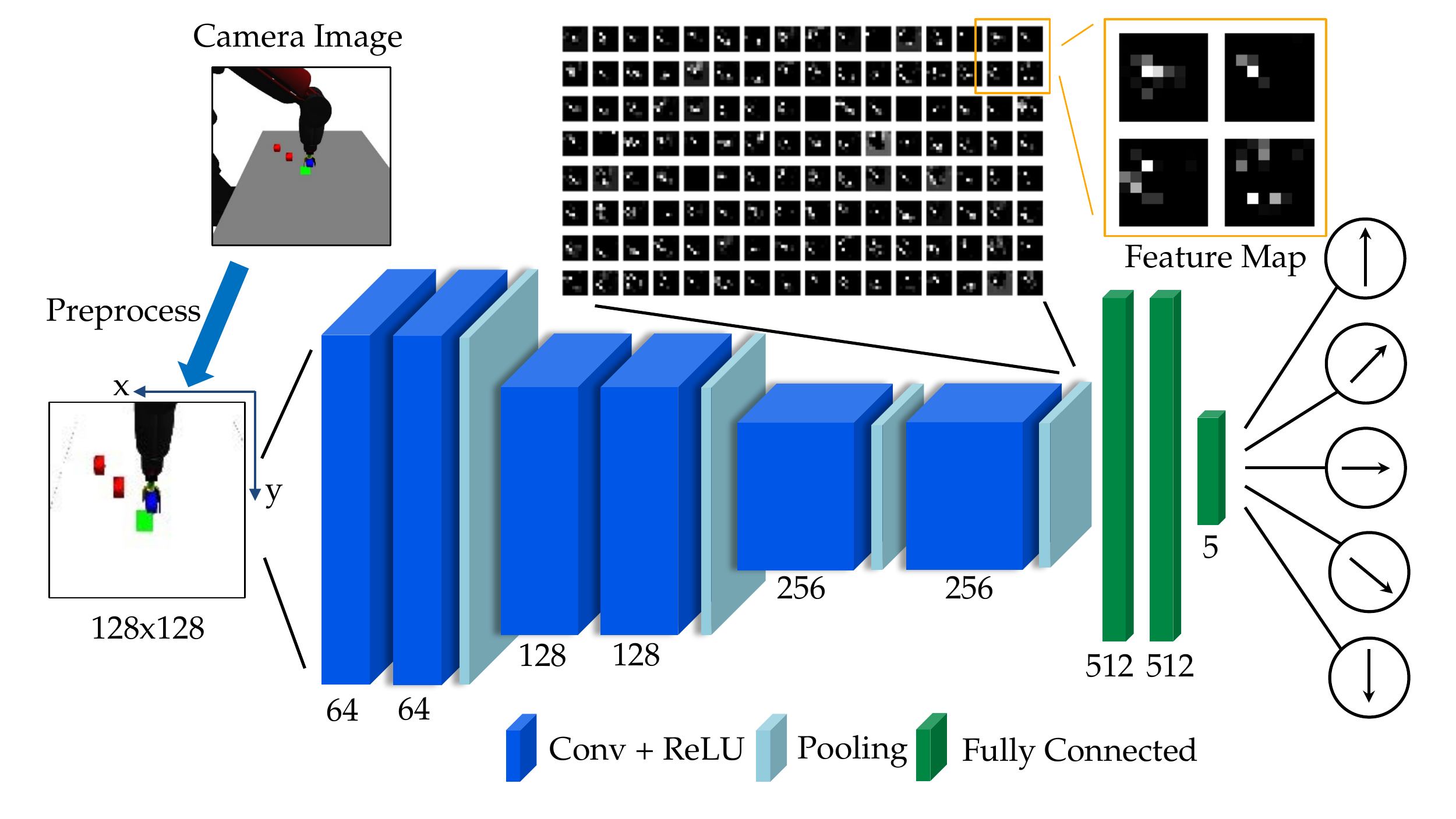}
\caption{We represent the action-value function with the deep convolutional neural network structure depicted here. The network computes Q-values for each action in parallel. The picture captured by the camera, the 128x128 image to be fed to the network and the feature map output by the convolutional part are shown.}
\label{fig:network_structure}
\vspace{-0.5cm}
\end{figure}

Our method is based on learning a deep $Q$-network \cite{mnih2015human} from experiences while using Gaussian potential fields \cite{Zhu06, Varava-RSS-17} to generate pertinent and informative examples during exploration.
% Below, we briefly describe these two concepts.

\subsection{Deep Q-Learning}
\label{sec:deep-q-learning}

Deep $Q$-learning considers tasks in which the agent interacts with the environment through a sequence of observations $\mathbf{x}_t$, actions $a_t$, and rewards $r_t$. The goal is to select actions that maximize cumulative reward. For this, the optimal state-action function ($Q$-function \cite{mnih2015human}),
\begin{multline}
Q^*(\mathbf{x}, a)
= \\
\max_\pi
\mathbb{E}
\left[ r_t + \gamma r_{t+1} + \gamma^2 r_{t+2} + \dots \mid \mathbf{x}_t = \mathbf{x}, a_t = a, \pi \right]
,
\end{multline}
is approximated by a deep convolutional neural network $Q(\mathbf{x}, a; \theta)$ with parameters $\theta$. This is the maximum sum of rewards discounted by $\gamma$ achieved by the policy $\pi$ after making observation $\mathbf{x}$ and taking action $a$.

%\subsubsection*{Training}

Representing the state-action function by a nonlinear function approximator can lead to instability and divergence \cite{tsitsiklis1997analysis}. These problems are usually addressed by \emph{experience replay} \cite{mnih2015human, mcclelland1995there, o2010play, lin1993reinforcement} and by training separate \emph{target} and \emph{primary networks}, with parameters $\theta^t$ and $\theta^p$ respectively, which are updated in different frequencies \cite{mnih2015human}.
% (to reduce correlation between and consecutive examples).
For this, previous experiences $e_t = (\mathbf{x}_t, a_t, r_t, \mathbf{x}_{t+1})$ from time steps $t$ are stored in a replay buffer $D$ to optimize the loss function,
\begin{multline}
\mathcal{L} (\theta^p)
=
\\
\mathop{\mathbb{E}}_{(\mathbf{x}, a, r, \mathbf{x}^\prime)}
\left[
\left(
r + \gamma \max_{a^\prime \in \mathcal{A}}
Q(\mathbf{x}^\prime, a^\prime; \theta^t)
-
Q(\mathbf{x}, a; \theta^p)
\right)^2
\right]
\label{eq:q-objective}
,
\end{multline}
for which $(\mathbf{x}, a, r, \mathbf{x}^\prime)$ is sampled from $D$ according to some distribution. The target network parameters are updated towards the primary network parameters upon a certain schedule.

Once the network is successfully trained, the greedy policy which selects the action with the maximal $Q$-value,
\begin{equation}
\pi(\mathbf{x})
=
\arg \max_{a \in \mathcal{A}} Q(\mathbf{x}, a; \theta)
\label{eq:greedy-policy}
,
\end{equation}
can be used to select actions to solve the task.

\subsection{Potential Fields}
\label{sec:potential-fields}

For planar navigation tasks, it is a common practice to model the effort or cost of passing through a point $\mathbf{p} \in \mathbb{R}^2$ by a potential field $U \colon \mathbb{R}^2 \to \mathbb{R}$ where higher potential means more effort required \cite{choset2005principles}. For identifying locally optimal motion directions $\theta \in [0, 2 \pi]$ at a point $\mathbf{p} \in \mathbb{R}^2$, we can consider the directional derivative $\nabla_{\mathbf{v}_\theta} U(\mathbf{p})$ along the vector $\mathbf{v}_\theta = [\sin \theta, \cos \theta]^\intercal$.
%
% \begin{equation}
% \nabla_{\mathbf{v}_\theta} U(\mathbf{p})
% =
% \mathbf{v}_\theta
% \cdot
% \frac{\partial U(\mathbf{p})}{\partial \mathbf{p}}
% ,
% \end{equation}
%
% which is defined by a dot product of the direction $ \mathbf{v}_\theta$ and the gradient vector where $|| \mathbf{v}_\theta || = 1$.

For simplicity, the potential field $U$ is often defined as a mixture of potential functions $U_i$, representing individual features of the environment,
\begin{equation}
U(\mathbf{p})
=
\frac{1}{N} \sum_{i = 0} ^{N} U_i(\mathbf{p})
.
\end{equation}
In \emph{Gaussian potential fields}, obstacles are modeled by the normal distribution function, $ U_i(\mathbf{p}) = \varphi(\mathbf{p}; \boldsymbol{\mu}, \boldsymbol{\Sigma})$, leading to a smooth potential surface. If the potential is independent for each dimension, i.e. the covariance matrix is diagonal, $\boldsymbol{\Sigma} = \mathrm{diag}(\sigma_x, \sigma_y)$, the potentials $U_i$ can be factorized,
\begin{equation}
U_i(\mathbf{p})
=
\varphi( \mathbf{p}_x; \boldsymbol{\mu}_x, \sigma_x)
\,
\varphi( \mathbf{p}_y; \boldsymbol{\mu}_y, \sigma_y)
,
\end{equation}
where subscript $x$ and $y$ are used to refer to dimensions one and two respectively. We use both, the normal distribution function $\varphi$ and the skew-normal distribution function $\varphi_\alpha$ with shape parameter $\alpha$ for modeling obstacles.

%\section{METHOD}
\section{LEARNING NONPREHENSILE REARRANGEMENT}
\label{sec:method}

%\direction{Describe the work in a way that others are able to re-implement it}
%\direction{Describe foundations if necessary}
%\direction{Give sufficient technical details}
%\direction{Include the underlying equations}
%\direction{Add figures to make description more easy to understand}
%\direction{Mention the advantages of the approach}
%\direction{Describe the complexity}

We learn nonprehensile rearrangement using $Q$-learning where the $Q$-function is approximated by a deep convolutional neural network. To train this network, we define rewards that model the task and alternate between collecting episodes of experiences and updating network parameters. Effective deep $Q$-learning requires both, informative and task-relevant experiences, and adequate utilization of past experiences. Below, we explain how we collect informative experiences by informed action sampling and how we utilize both failure and success in learning by sampling the replay buffer. The process is summarized in Alg.~\ref{alg:learning}.

%\todo{Give abstract summary and overview.}

\begin{algorithm}[htb]
\caption{Learning Architecture}
\label{alg:learning}

\begin{algorithmic}[1]
\State Randomly initialize primary and target networks $\theta^p = \theta^t$
\State Initialize experience buffer $D$
\For{episode $k = 1, 2 \dots, K$}
\For{time step $ t = 1, 2,  \dots $ \textbf{until} termination}
\If{ with probability $P_{\text{exploit}}$ }
% EXPLOIT
\State $a_t \gets \arg \max_{a \in \mathcal{A}} Q(\mathbf{x}, a; \theta^t)$
\Else
% EXPLORE
\State Sample action $a_t \sim P_{\mathcal{A}}(a \mid \mathbf{p}^{\text{tool}})$
\Comment{Sec.~\ref{sec:exploration}}
\EndIf
\State Execute $a_t$
\State Get experience $e_t = (\mathbf{x}_t, a_t, r_t, \mathbf{x}_{t+1})$
\EndFor
\State Update $D$ according to policy \Comment{Sec.~\ref{sec:buffer-policy}}
\State Sample experiences $e \sim \mathrm{Uniform}(D)$
\State Update $\theta^p$ and $\theta^t$ according to policy \Comment{Sec.~\ref{sec:update-policy}}
\EndFor
\end{algorithmic}
\end{algorithm}

\subsection{Network Structure}
\label{sec:network-structure}

We define a deep convolutional neural network that computes the action-value function for each action $a \in \mathcal{A}$ in parallel. The input of the network is one observation $\mathbf{x}$ with $128 \times 128$ RGB pixels and the output is the $Q$-values for actions. As seen in Fig.~\ref{fig:network_structure}, there is a convolutional part for learning a low-dimensional representation followed by a fully connected part for mapping to action values. The convolutional part consists of six convolutional layers with Rectified Linear Unit (ReLU) as  activation function to extract the feature map, and four max pooling layers to reduce the size of the output. This network structure is instantiated twice for the target network and the primary network respectively.

% learning
%there are two same network, one is primary network for direct parameters updating from training data and one is target network \cite{mnih2013playing} for Q-value calculating, which updates parameters slowly and improves the stability of the system.

\subsection{Reward}
\label{sec:reward}

In reinforcement learning, the reward implicitly specifies what the agent is encouraged to do. Therefore, it is important that the reward models the task correctly. We want to relocate the manipulation object to the target location by moving the manipulation tool but without obstacle collision.  For this, we define the reward given an episode of experiences of length $T$ using three components $r_t^{\text{tool}}$, $r_t^{\text{man}}$, and $r_t^{\text{term}}$.
%This requires that the tool moves towards the object and can be modeled the formula,
%
The first component,
\begin{equation}
r_t^{\text{tool}}
=
(\left\Vert \mathbf{p}^{\text{tool}}_{t-1} - \mathbf{p}^{\text{man}}_{t-1} \right\Vert
-
\left\Vert \mathbf{p}^{\text{tool}}_t - \mathbf{p}^{\text{man}}_t \right\Vert)/d_a
,
\end{equation}
increases when the tool and the manipulation object get closer.
The second component,
\begin{equation}
r_t^{\text{target}}
=
(\left\Vert \mathbf{p}^{\text{man}}_{t-1} - \mathbf{p}^{\text{target}}_{t-1} \right\Vert
-
\left\Vert \mathbf{p}^{\text{man}}_t - \mathbf{p}^{\text{target}}_t \right\Vert)/d_a
,
\end{equation}
increases when the manipulation object and the target location get closer.
Finally, we have to capture success or failure which only occurs at the end of the episode at time step $T$. In case of success, the manipulation object reaches the target location, $\left\Vert \mathbf{p}^{\text{man}}_T - \mathbf{p}^{\text{target}}_T \right\Vert < \epsilon^{\text{suc}}$. The episode is terminated with failure when too many steps have been taken, obstacles are moved (collision), or the tool moves out of the work-surface. We model these conditions by the following terminal reward,
\begin{equation}
r_T^{\text{term}}
=
\begin{cases}
\phantom{-}1,
&  \Vert \mathbf{p}^{\text{man}}_T - \mathbf{p}^{\text{target}}_T \Vert <  \epsilon^{\text{suc}}
\\
-1,
& \Vert \mathbf{p}^{\text{obs}, i}_T - \mathbf{p}^{\text{obs}, i}_{0} \Vert > \epsilon^{\text{fail}}
%\sum_i  \left\Vert \mathbf{p}^{\text{obs}, i}_t - \mathbf{p}^{\text{obs}, i}_t \right\Vert > \epsilon^{\text{move}}
 \\
-1,
&  \mathbf{p}_T^{\text{tool}} \text{ out of work-surface}
\\
-1,
& T = n^{\text{steps}} \quad \text{(timeout)}
\end{cases}
\, ,
\end{equation}
which is $0$ for all steps $t < T$. The last reward captures the main essence of the task but is an infrequent experience. 

%We assign credit for success and failure to all previous experiences of the episode by defining,
%
%\begin{equation}
%r_t^{\text{term}}
%=
%\gamma^{T-t} r_T^{\text{term}}
%,
%\quad
%\forall t < T
%.
%\end{equation}

All three rewards defined above are combined in a weighted sum with $\alpha_1, \alpha_2$ and $\alpha_3$ being the weighing factor:
\begin{equation}
r_t
=
\alpha_1 r_t^{\text{tool}}
% r_t^{\text{tool}}
+
\alpha_2 r_t^{\text{target}}
% r_t^{\text{target}}
+
\alpha_3 r_t^{\text{term}}
% r_t^{\text{term}}
,
\end{equation}
to form our reward feedback.

\subsection{Heuristic Exploration with Informed Action Sampling}
\label{sec:exploration}

\begin{figure}[tbh]
\centering
\includegraphics[width=0.9\columnwidth]{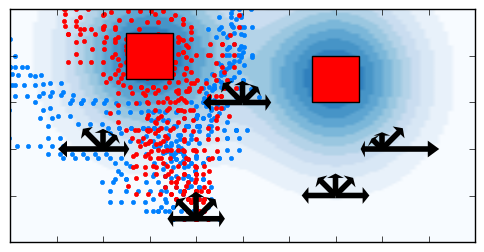}
\caption{For exploration, we model the environment with obstacles (red squares) as a Gaussian potential field and sample actions according to local potential changes. This process selects actions leading away from obstacles more frequently than actions leading towards obstacles. Arrow length indicates action probability. For illustration, we sample actions uniformly (red) and according to our distribution (blue) starting from the same position. Red paths lead to collisions more often than blue paths.}

%\label{fig:contour}
%\framebox{
%\includegraphics[width=0.85\columnwidth]{Image/direction.png}
%}
%\caption{Five discrete actions divide the moving direction to five sector areas. For every action we calculate a probability according to the position of object and obstacles}
%\label{fig:direction}

%\todo{These two figures could be combined into one to illustrate the concept more clearly. Also a figure that shows the projection of the potential field on the circle's radius could be interesting.}
\label{sec:potential-example}
\end{figure}

Reinforcement learning for our rearrangement task is challenging because exploration can lead to preemptive termination of an episode. For example, when the manipulation tool is close to obstacles, uniformly sampling the next action is often a poor choice leading to collisions as seen as red dots in Fig.~\ref{sec:potential-example}. Doing so nevertheless ultimately results in an unbalanced training set dominated by unsuccessful episodes. When exploring, we therefore aim at selecting actions that are unlikely to prematurely terminate the episode due to obstacle collisions as a means to collect informative samples for the dataset similar to \cite{Zhu06, Varava-RSS-17}.

\subsubsection{Action Sampling}
We are interested in a complete heuristic for exploration that does not preclude certain types of experiences but want to sample actions such that collisions are infrequent. For this reason, we model the environment by a potential field $U$, as described in Sec.~\ref{sec:potential-fields}, and sample exploration actions from a distribution, $a \sim P_{\mathcal{A}}$, which depends on local potential change. This results in a lower frequency of actions moving the tool close to obstacles, which increases potential, and higher frequency of actions moving the tool into an obstacle-free region, which decreases potential as illustrated by blue dots in Fig.~\ref{sec:potential-example}.

We model $P_{\mathcal{A}}$ by discretizing the space of motion directions $\theta$ between $-\frac{1}{8}\pi$ and $\frac{9}{8}\pi$ into five intervals, as shown in Fig.~\ref{fig:actions}, resulting in sectors $I_a$ centered around each action's motion direction. To compute an action's probability $P_{\mathcal{A}}(a \mid \mathbf{p})$ at a point in the environment $\mathbf{p} \in \mathbb{R}^2$, we first integrate potential change in $U$ at position $\mathbf{p}$ over the angle interval $I_a$,
\begin{equation}
\Delta(a, \mathbf{p})
=
\int_{\theta \in I_a} \nabla_{\mathbf{v}_\theta} U(\mathbf{p}) \mathrm{d} \theta
,
\end{equation}
for each action $a \in \mathcal{A}$. Based on the potential changes $\Delta(a, \mathbf{p})$, we formulate the distribution $P_{\mathcal{A}}$ using a normalized exponential function,
\begin{equation}
P_{\mathcal{A}}(a \mid \mathbf{p})
=
\frac{\exp (-\Delta(a, \mathbf{p})) }{\sum_{a^\prime \in \mathcal{A}}\exp ( -\Delta(a^\prime, \mathbf{p}))}
\label{eq:action-distribution}
,
\end{equation}
which assigns higher probability to larger instantaneous reduction of potential.

% \todo{Indicate $I_a$ in Fig. 2. Show the tool in Fig. 2. Show axis in Fig. 2 and Fig. 3. }

\subsubsection{Environment Model}

For sampling actions according to Eq.~\eqref{eq:action-distribution}, we assume that the tool is oriented along the $Y$-axis and define obstacle potentials $U_i$ consisting of two factors for each obstacle with position $\mathbf{p}^{\text{obs}, i}$,
\begin{equation}
U_i( \mathbf{p} )
=
\varphi(\mathbf{p}_x; \mathbf{p}^{\text{obs}, i}_x, \sigma_i)
\,
\varphi_\alpha( \mathbf{p}_y; \mathbf{p}^{\text{obs}, i}_y, \sigma_i)
,
\end{equation}
where we use the notation from Sec.~\ref{sec:potential-fields}. Skewing the potential along the $Y$-axis makes the potential steeper when the tool is before the obstacle leading to stronger emphasize on avoiding collisions.

\subsection{Experience Replay and Network Updates}
\label{sec:replay-updates}
%-- Infrequent feedback about success failure (delayed reward) in episodic task
%-- Exploration is difficult because most action sequences are stupid (large space of sequences) (Curse of dimensionality)

The stability-plasticity dilemma and correlation of experience \cite{mnih2015human} in deep $Q$-learning are usually addressed by uniformly sampling experiences from the replay buffer $D$ of  previous experiences \cite{mnih2015human, lillicrap2015continuous, schaul2015prioritized, adam2012experience} for training. However, until the task has been sufficiently learned the majority of experiences would come from failed episodes, e.g., the manipulation tool did not catch the object, the motion caused collisions or the motion did not lead to the goal region. In our experience, this leads to slow learning in our task.

% missing the delayed reward signal from placing the manipulation object at the goal location.

For effective training, sampled experiences need to be informative and representative, which in our experience means that they should come from successful \emph{and} failing episodes in equal shares. Additionally, when learning a task with high-dimensional observations, not all experiences can be collected in the buffer $D$ and adding new experiences displaces older ones. Therefore, we propose a policy for data sampling and storing and a policy for network updates.

\subsubsection{Replay Buffer Policy}
\label{sec:buffer-policy}

The overall goal is to avoid over-representing failed or successful episodes in training data. For this, we store experiences in $D$ according to variable probabilities $P_{\text{store}}$. If the ratio of successful experiences in the buffer is far away from $50\%$, e.g. less than $30\%$, we use a higher storing probability $P_{\text{store}}$ for successful experiences than for failing experiences. If ratio is more than $70\%$, we do the opposite. If the buffer is full, the oldest experience is displaced by the newly added experience.
% In the limit, this makes sure that at least 50\% of the experiences come from successful episode.

% \todo{Flow chart?}

\subsubsection{Network Update Policy}
\label{sec:update-policy}

Updating the network parameters with experiences from a dataset biased towards failing episodes leads to poor performance on the task. Therefore, we update the network according to the dataset's condition. If the ratio of success experiences deviates into any direction from $50\%$, we slow down the network update in terms of the deviation magnitude. The schedule based on the ratio of success experiences $r^{\text{succ}}$ shown below realizes this concept:
\begin{multline}
\text{Update action:}
\notag
\\
\begin{cases}
\text{Update with small probability},
&
\epsilon_1 < |r^{\text{succ}}-0.5|
\\
\text{Update once},
&
\epsilon_2 \leq |r^{\text{succ}}-0.5| < \epsilon_1
\\
\text{Update multiple times},
&
|r^{\text{succ}}-0.5| < \epsilon_2
\end{cases}
\label{eq:update-schedule}
\end{multline}
where $\epsilon_{i}$ are the update control points.
Whenever we update the primary network, we update the target network parameters $\theta^t$ according to the primary network's parameters $\theta^p$ using a low learning rate of $0.001$,
\begin{equation}
\theta^t \gets 0.999 \, \theta^t + 0.001 \, \theta^p
,
\end{equation}
which leads to slow adaptation but increases learning stability.

% \todo{Flow chart?}

%\begin{figure}
% \centering
%       \subfigure[Successful episodes dominant]{
%    \label{buffer1}
%    \includegraphics[width=0.2\textwidth]{Image/buffer1.png}
%    }
%    \subfigure[Failing episodes dominant]{
%    \label{buffer2}
%    \includegraphics[width=0.2\textwidth]{Image/buffer2.png}
%    }
%    \subfigure[Balanced buffer]{
%    \label{buffer3}
%    \includegraphics[width=0.2\textwidth]{Image/buffer3.png}
%    }
%  \caption{Replay buffer is composed of successful episodes denoted in blue and failing episodes denoted in green. (a) shows the occasion where successful episodes dominate while (b) shows the opposite occasion; (c) shows the balanced occasion we want.}
%  \label{buffer}
%\todo{Can this figure be changed to show the decision making process for when to save or not save experiences?}
%\end{figure}

% !TEX root =  ../main.tex

\section{EXPERIMENTS}
\label{sec:experiments}

% \direction{Explain why you make the individual experiment}
% \direction{Motivate simulation and real-robot experiments}
% \direction{Give a detailed explanation of the individual experiments}
% \direction{Eventually, use graphs and tables to summarize your experiment}
% \direction{Compare your approach to alternative ones}
% \direction{Perform statistical tests indicating that your approach is significantly better than alternatives}

In this section, we present experiment setup, data collection, model training and evaluation. We quantitatively evaluate the DQN trained using our approach to show that it can handle the given task with high success rate. Additionally, we provide qualitatively examples that demonstrate how our approach reacts to sudden changes from external influences and how it generalizes to slight changes of physical properties.

\subsection{Experiment Platform and Setup}

The experiments are conducted with a Baxter robot in a simulated virtual environment using Gazebo \cite{Koenig14}. The simulation considers physical properties such as mass, friction, and velocities, but these are not known to the robot. A customized manipulation tool is mounted on the left hand of Baxter as seen in Fig.~\ref{fig:actions}. The robot only controls its left arm to interact with the environment. The manipulation object and obstacle objects are represented by cube-shaped objects. For perception, we simulate a fixed camera beside the robot as shown in Fig.~\ref{fig:first}. We define the work-surface to be 30 by 50 cm. The system parameters are empirically determined in terms of both the performance and our computation resource limits as listed in Table \ref{tab:parameter}.

\begin{table}[h]
\centering
\caption{System Parameters}
\begin{tabular}{r c c}
\toprule
Parameter & Notation & Value \\
\midrule
\rowcolor[gray]{0.95} 
Primary-Net Learning Rate & $\eta$ & $10^{-4}$\\
Replay Buffer Size & $|D|$ & $200,000$\\
\rowcolor[gray]{0.95} 
Discount Factor & $\gamma$ & $0.99$\\
Episode Limit & $n^{\text{steps}}$ & $150$\\
\rowcolor[gray]{0.95} 
Reward Weights&$\alpha_1,~\alpha_2,~\alpha_3$ & $0.1, 0.2, 1$\\
Update Policy& $\epsilon_1,~\epsilon_2$ & $0.4, 0.1$\\
\rowcolor[gray]{0.95} 
$\varepsilon$-greedy & $K_1,~K_2$ & $200, 5000$\\
Action Scale & $d_a$ & 1cm\\
\bottomrule
\end{tabular}
\label{tab:parameter}
\vspace{-0.4cm}
\end{table}

% \begin{figure*}[htb]
% \centering
% \input{Image/buffer.tex}
% \input{Image/suc_rate.tex}
% \input{Image/step_suc.tex}
%  \input{Image/obstacle234.tex}
% 
% \caption{(a) The ratio of success episodes in the replay buffer. BC: Replay Buffer Control. IAS: Informed Action Sampling. (b) The success rate against the number of experienced episodes. (c) The average number of actions taken to accomplish a random task. (d) The success rate in test scenes with $2$, $3$ and $4$ random obstacles.}
% \label{fig:quantitative}
% \end{figure*}

\begin{figure*}
  \includegraphics{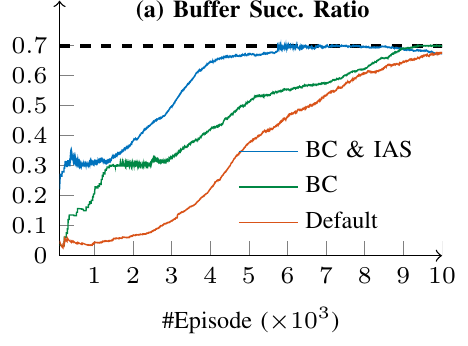}
  \includegraphics{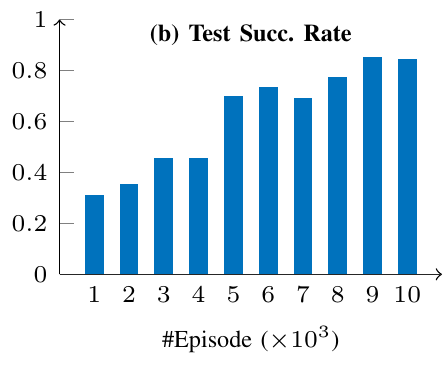}
  \includegraphics{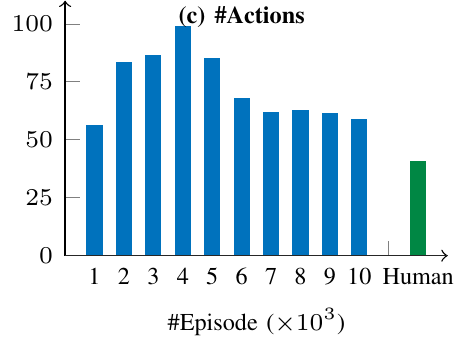}
  \includegraphics{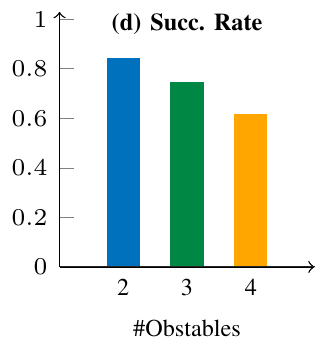}
  \caption{(a) The ratio of success episodes in the replay buffer. BC: Replay Buffer Control. IAS: Informed Action Sampling. (b) The success rate against the number of experienced episodes. (c) The average number of actions taken to accomplish a random task. (d) The success rate in test scenes with $2$, $3$ and $4$ random obstacles.}
\label{fig:quantitative}
\vspace{-0.4cm}
\end{figure*}

\subsection{Data Collection}

For each training episode, we initialize the robot in the starting pose and randomly place the manipulation object in front of the manipulation tool. The number of obstacles is fixed to $2$ for data collection. The obstacles are placed randomly while at least one obstacle is directly placed between the manipulation object and the target location making obstacle avoidance necessary. We set the maximal episode length to $n^{\text{steps}}$ and proceed according to Alg.~\ref{alg:learning} to select actions and update the network.

Exploiting with a poor initial training policy rarely leads to successful episodes. Therefore, we tradeoff between \emph{exploration} and \emph{exploitation} using an $\epsilon$-greedy training schedule with three phases \cite{mnih2015human}: At the beginning of training, the convolutional part of the network is not well trained, so 1) we employ \emph{only exploration} (Sec.~\ref{sec:exploration}) for $K_1$ episodes to train state perception; 2) after this phase, we increase the exploitation probability for each episode until episode $K_2$ and 3) thereafter, we only train with exploitation for learning the state-action function. This is summarized below as the the exploration probability $P_{\text{exploit}}$ for episode number $k$,
\begin{equation}
P_{\text{exploit}}
=
\begin{cases}
0,
&
0 < k \leq K_1
\\
\tau k,
&
K_1 < k \leq K_2
\\
1,
&
k > K_2
\end{cases}
\label{eq:epsilon-greedy}
,
\end{equation}
where $\tau \in \mathbb{R}^+$ is a factor which controls the probability $P_{\text{exploit}}$ to linearly increase from $0$ to $1$ in the corresponding range.

\subsection{Network Training}

While collecting experiences as aforementioned, we train the deep $Q$-network \emph{de novo} in terms of objective function Eq.~\eqref{eq:q-objective} with the Adam optimizer \cite{kingma2014adam}. The mini-batch size is set to 32. In order to evaluate the proposed approach, we train the network using $3$ different configurations: 1) The network is trained using the replay buffer control (Sec.~\ref{sec:replay-updates}) and the informed action sampling (Sec~\ref{sec:exploration}). 2) The network is trained using only buffer control. 3) The network is trained without any of proposed methods. The training process took approximately 600k actions during which 10k episodes were collected for each of the $3$ configurations. The training hardware is a single Nvidia GeForce GTX 1080 Ti GPU. More than 90\% of training time is spent on simulation.

\subsection{Quantitative Experiments}

\subsubsection{Reply Buffer Control and Informed Action Sampling}
%As explained in Sec.~\ref{sec:exploration}, we employ a potential field for informed action sampling, which can help to select collision-free actions during exploration to avoid preemptive termination of an episode, so as to facilitate the collection of successful experiences for network training. Additionally, we adaptively control the ratio between success and failure experiences in the replay buffer as in Sec.~\ref{sec:replay-updates}, in order to avoid biased training.

For evaluating the effectiveness of these two techniques, we record the ratio of success episodes in the reply buffer during the training process of the aforementioned $3$ training configurations. As shown in Fig.~\ref{fig:quantitative}(a), the default configuration performs poorly until $7,000$ episodes to achieve $50\%$, which has been achieved at the episode $5,000$ by adding the buffer control. By additionally applying informed action sampling, the buffer achieves a balanced share already at the episode $3,000$. This result clearly shows the effectiveness of our proposed methods. Furthermore, as explained below, it is crucial to collect sufficient success experiences for training, since it significantly affects the training results.

\begin{figure*}
\centering
  \begin{subfigure}[b]{0.35\columnwidth}
    \includegraphics[width=\columnwidth, trim={3cm 8cm 5cm 0cm}, clip]{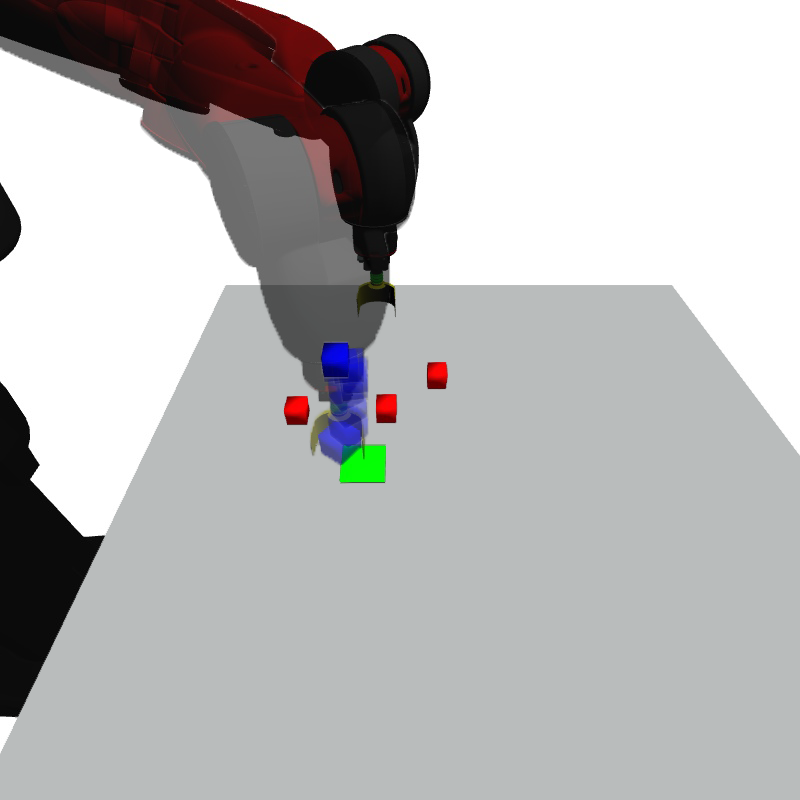}
    \caption{3 obstacles}
    \label{fig:3_obstacle}
  \end{subfigure}
  \begin{subfigure}[b]{0.35\columnwidth}
    \includegraphics[width=\columnwidth, trim={3cm 8cm 5cm 0cm}, clip]{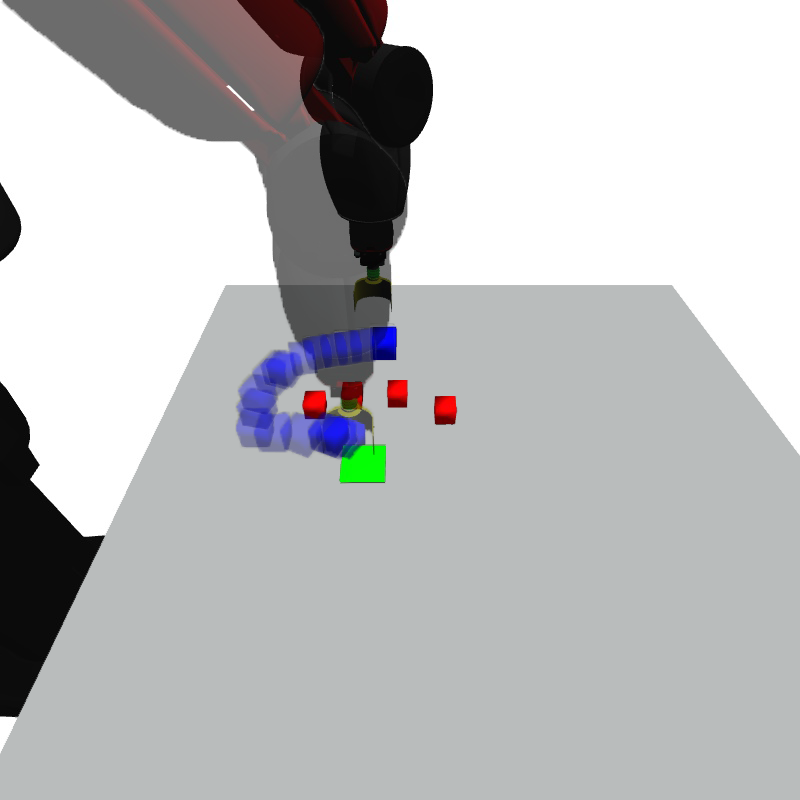}
    \caption{4 obstacles}
    \label{fig:4_obstacle}
  \end{subfigure}
  \begin{subfigure}[b]{0.35\columnwidth}
    \includegraphics[width=\columnwidth, trim={3cm 8cm 5cm 0cm}, clip]{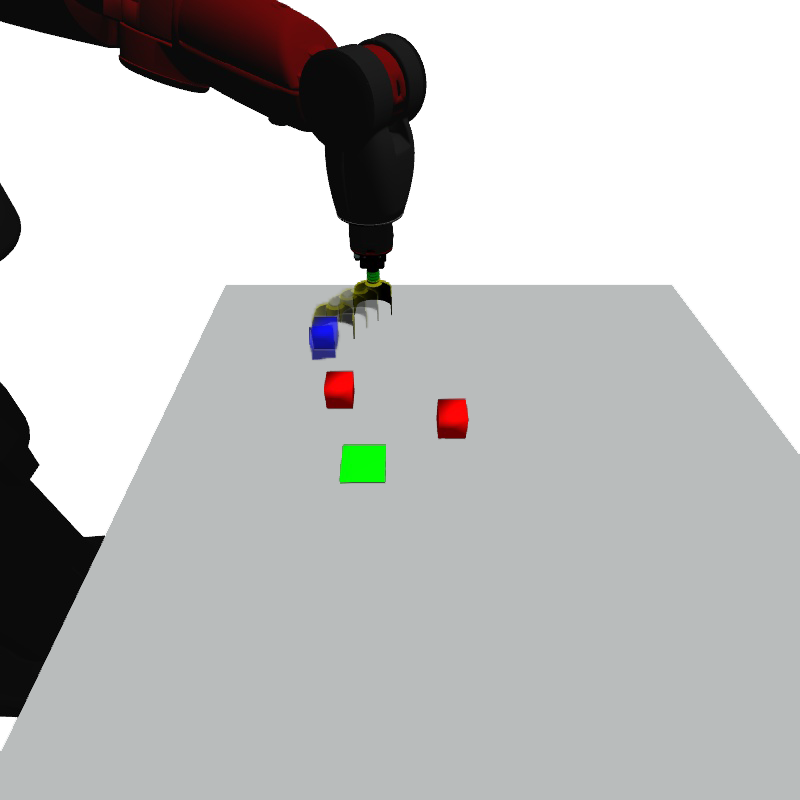}
    \caption{Before object moved}
    \label{fig:object_move1}
  \end{subfigure}
  \begin{subfigure}[b]{0.35\columnwidth}
    \includegraphics[width=\columnwidth, trim={3cm 8cm 5cm 0cm}, clip]{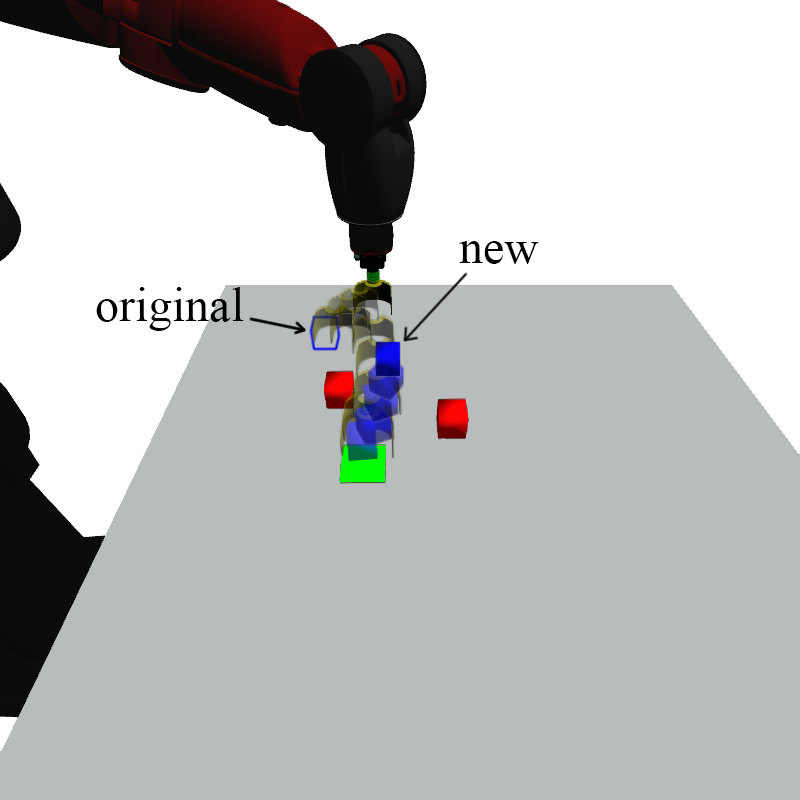}
    \caption{Object suddenly moved}
    \label{fig:object_move2}
  \end{subfigure}
  \begin{subfigure}[b]{0.35\columnwidth}
    \includegraphics[width=\columnwidth, trim={3cm 8cm 5cm 0cm}, clip]{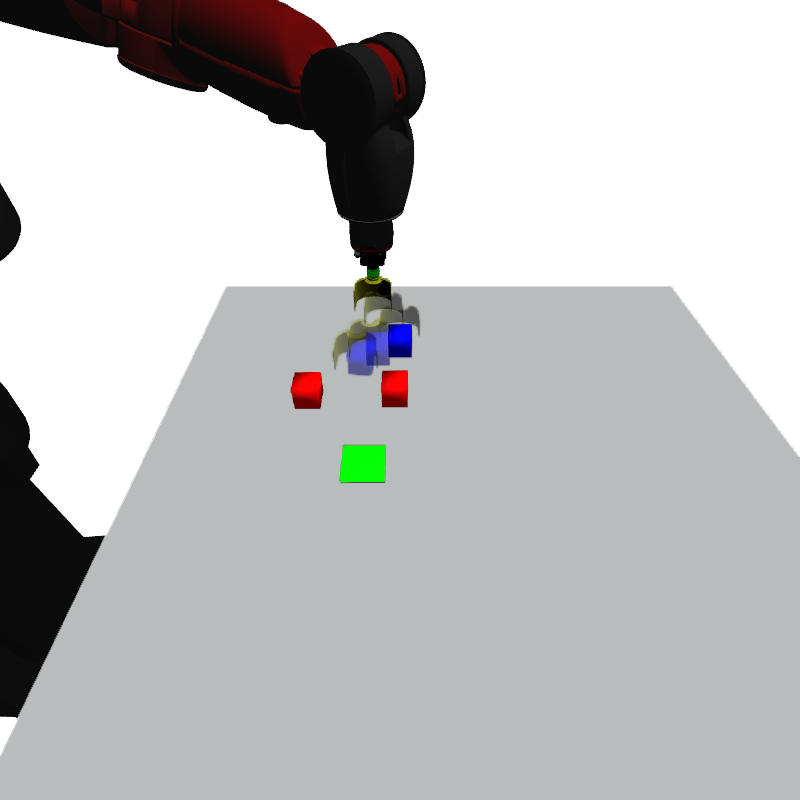}
    \caption{Before obstacle moved}
    \label{fig:obstacle_move1}
  \end{subfigure}
  \begin{subfigure}[b]{0.35\columnwidth}
    \includegraphics[width=\columnwidth, trim={3cm 8cm 5cm 0cm}, clip]{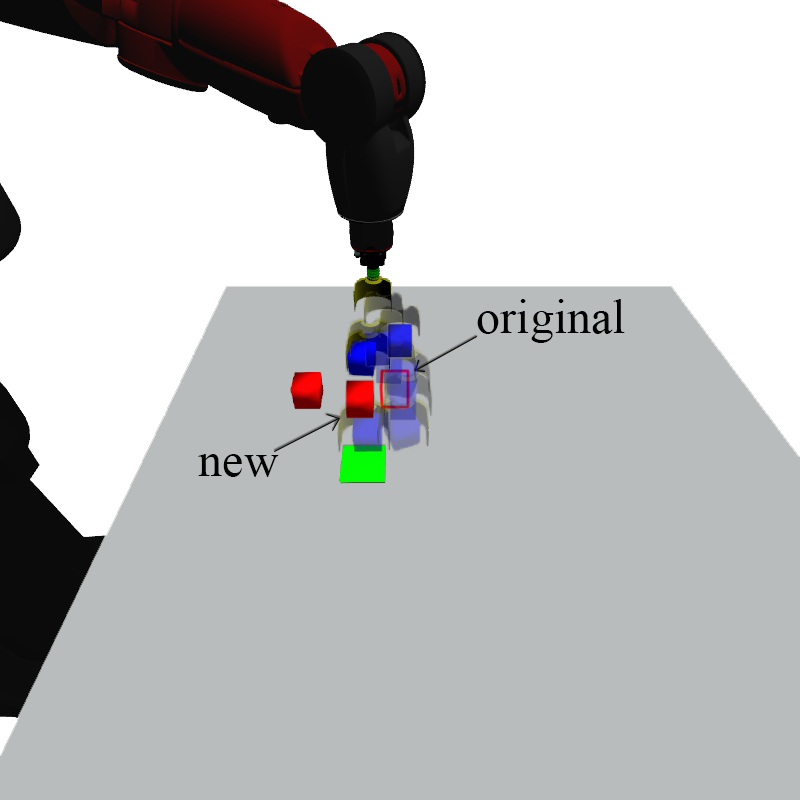}
    \caption{Obstacle suddenly moved}
    \label{fig:obstacle_move2}
  \end{subfigure}
  \begin{subfigure}[b]{0.35\columnwidth}
    \includegraphics[width=\columnwidth, trim={3cm 8cm 5cm 0cm}, clip]{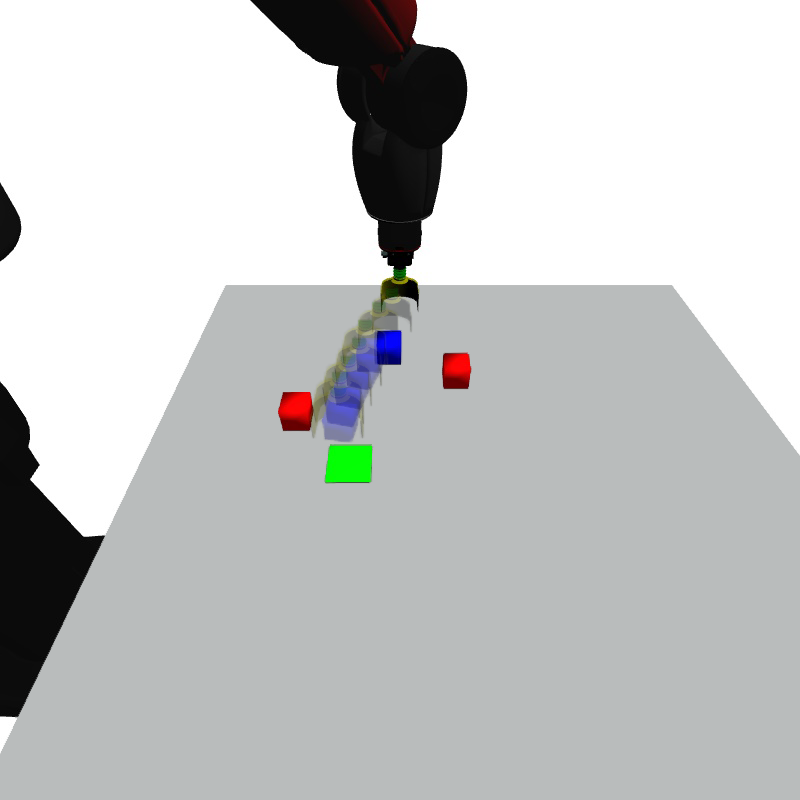}
    \caption{Before target moved}
    \label{fig:target_move1}
  \end{subfigure}
  \begin{subfigure}[b]{0.35\columnwidth}
    \includegraphics[width=\columnwidth, trim={3cm 8cm 5cm 0cm}, clip]{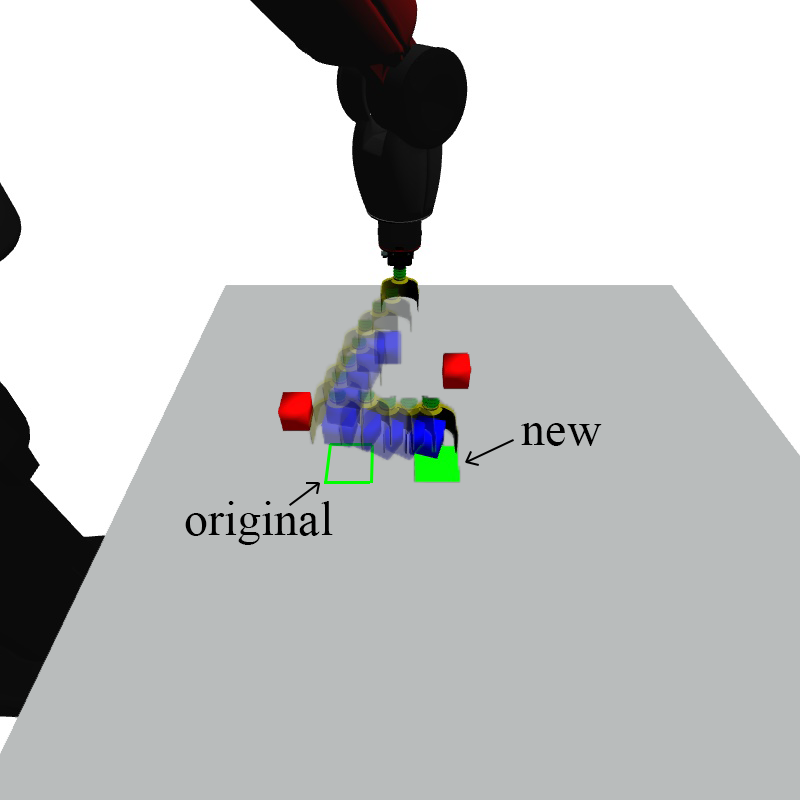}
    \caption{Target suddenly moved}
    \label{fig:target_move2}
  \end{subfigure}
  \begin{subfigure}[b]{0.35\columnwidth}
    \includegraphics[width=\columnwidth, trim={3cm 8cm 5cm 0cm}, clip]{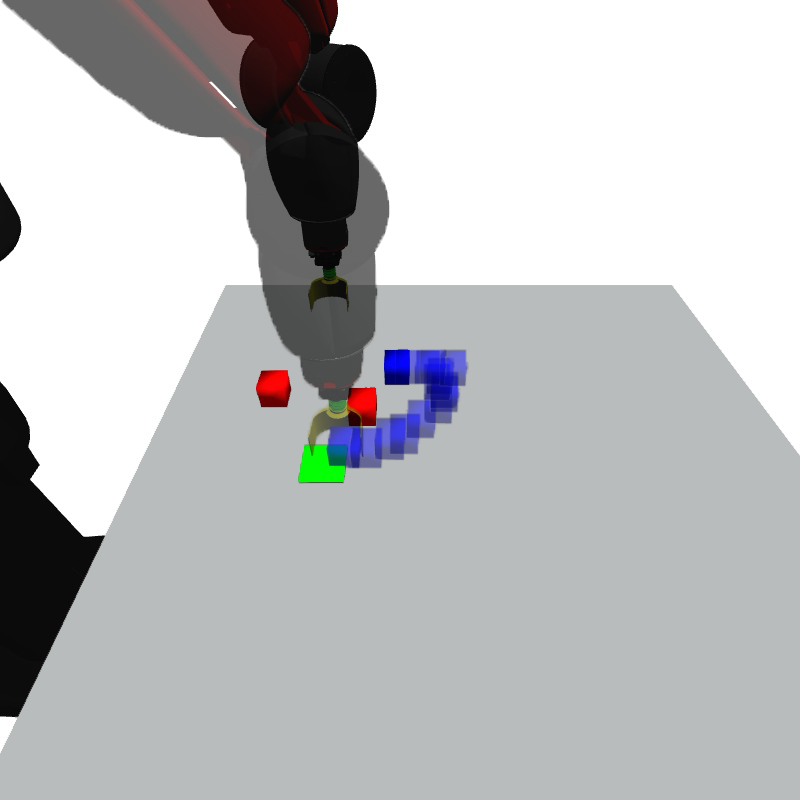}
    \caption{Low-friction}
    \label{fig:friction_change}
  \end{subfigure}
  \begin{subfigure}[b]{0.35\columnwidth}
    \includegraphics[width=\columnwidth, trim={3cm 8cm 5cm 0cm}, clip]{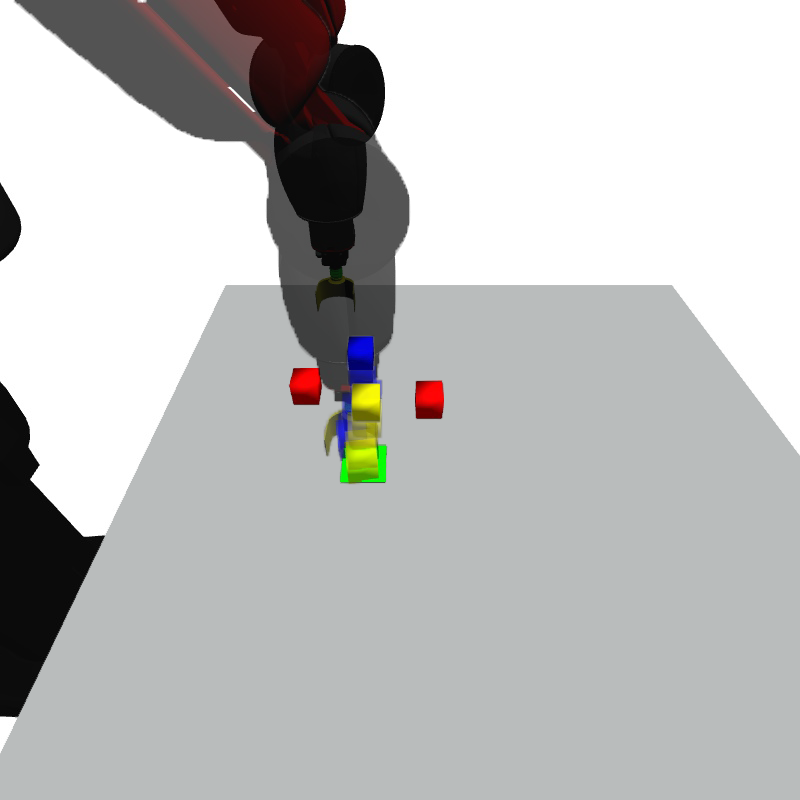}
    \caption{Distraction object}
    \label{fig:yellow}
  \end{subfigure}
  %

%   \subfigure[Phisical Properties Change]{
%   \label{fig:friction_change}
%   \includegraphics[width=0.2\textwidth,  trim={3cm 10cm 8cm 5cm}, clip]{Image/friction_change.png}
%   }
%   \subfigure[Disturbing Block]{
%   \label{fig:yellow}
%   \includegraphics[width=0.2\textwidth,  trim={3cm 10cm 8cm 5cm}, clip]{Image/yellow.png}
%   }
%
% \caption{Qualitative experiments to examine the robustness of the network. Object moving experiment is shown in (a)(b). In (a) the robot tries to move close to the object. We move the object to a new position suddenly. Then the robot realizes it and moves instantly towards the new position. (c)(d) shows how the robot avoids the obstacles. In (c) the robot wants to pass throuth the way between the two obstacles. But then we change the position of one obstacle to block off the road. In this case the robot changes its moving direction immediately and bypasses the obstacle. And target moving occasion is also shown as (e)(f). (g)(h) show some environments change experiments to examine the robustness of the network. Phisical properties's change adds the difficulty of the task and robot needs more steps to accomplish it while the yellow block does not influence the work.}
\caption{Qualitative experiments to investigate the robustness of the network. (a-b) Example executions when $3$ or $4$ obstacles were randomly positioned. (c-h) Reactive path re-planning when the manipulation object, obstacles or the target positions were suddenly moved. (i) Reactive action planning in a low-friction environment. (j) Example execution when a distraction object (yellow) was involved.}
\label{fig:qualitative}
\vspace{-0.4cm}
\end{figure*}

\subsubsection{General Performance}

During training, we save the network parameters once every $1,000$ episodes and evaluate its performance using $300$ random scenes. As shown in Fig.~\ref{fig:quantitative}(b), the success rate increases while the network experienced more episodes and is stable and converged around the episode $10,000$. We can observe a rapid increment at the episode $5,000$ where the success episodes in the reply buffer reached the upper limit of $70\%$. This implies that sufficient success experiences in the reply buffer is crucial for increasing the network's performance. Finally, we achieve a success rate of $85\%$ indicating that the learned network can effectively handle the task of nonprehensile rearrangement.

\subsubsection{Action Effectiveness}

For the training process of the paragraph above, Fig.~\ref{fig:quantitative}(c) shows the number of actions needed to complete a random task. It can be observed that less actions are needed in the beginning of training. This is because in the beginning the network only succeeds in very simple scenes which do not require many actions. After experiencing $4,000$ episodes, the number of actions starts to decrease since the network has further optimized the reward outputs to make the actions more effective.

Additionally, we involve a human subject to be tested with the same input as the robot to make action decisions by pressing arrow keys to control the end-effector in the 2d workspace. The result in Fig.~\ref{fig:quantitative}(c) shows that the human performed a little better in terms of the number of actions. This is because our network is more conservative than the human in collision avoidance and tends to keep away from obstacles. However, this also shows that the effectiveness of our network is comparable to a human as it does not take many more actions to achieve the same tasks.

\subsubsection{Number of Obstacles}

Although we train the network using only $2$ random obstacles, we test it using also $3$ and $4$ obstacles in $300$ random scenes. As shown in Fig.~\ref{fig:quantitative}(d), the performance deteriorates when more obstacles are involved. However, the network is still able to handle most of the scenes. Example solutions generated by our network are shown in Fig.~\ref{fig:qualitative}(a-b). We interpret this as that the network learns not only global features to find the path of moving from the start position to the target position, but also local features to avoid collisions. We note that the failures in scenes with $3$ and $4$ obstacles can happen sometimes due to the target being fully blocked by randomly placed obstacles which do not allow for completing the task.

\subsection{Qualitative Experiments}

One of the most important advantages of a learned policy over a classic physics-based planning algorithm is that the final behavior naturally reacts to unexpected changes in the environment without the need of explicit re-planning. Below, we test robustness of our approach by moving objects and adding distractors during execution, as well as setting the friction coefficient different from training.
 
% The advantage of the real time end-to-end planning is its ability to deal with emergency and complex environment. So we try to test some situations beyond the normal running.
%
\subsubsection{Object Sliding, Obstacles Moving and Target Moving}

%During the execution of a rearrangement task, the positions of objects in the environment can suddenly change due to external influence. In such cases, classic planning algorithms would  fail or have to re-plan from scratch to resume execution. However, since our end-to-end network takes only the current image as the input, it can seamlessly react to those situations to continue the execution.

As shown in Fig.~\ref{fig:qualitative}(c-d), while the manipulation tool is approaching the manipulation object, we suddenly change the position of the manipulation object. Still, our approach can finish the task. Additionally, in Fig.~\ref{fig:qualitative}(e-f), we suddenly change the position of one of the obstacles to block the direct path. Again, our approach completes the task. Moreover, as seen in Fig.~\ref{fig:qualitative}(g-h), we suddenly change the target position when the robot is just about to complete the task. Here, our approach reaches the new target position.

\subsubsection{Low-friction Environment}

As another test, we significantly decrease the friction coefficient between the manipulation object and the table surface, such that the object will slide to some direction after each action. In Fig.~\ref{fig:qualitative}(i), we can see that although the object path is jittering during the execution, the approach still completes the task. This example is also presented in the complementary video.

\subsubsection{Distraction}

As shown in Fig.~\ref{fig:qualitative}(j), when there is an distraction object (yellow) in the environment, the behavior is not affected by it and can still complete the task. However, it is worthwhile to note that the distraction object is pushed. This indicates that the network focuses on the relevant information from the inputs, but that it does not guarantee collision-free manipulation with new, unknown objects.

\section{CONCLUSION}
\label{sec:conclusion}
% \direction{Again describe the approach presented in this paper}
% \direction{Again mention the advantages and what is novel compared to previous approaches}
% \direction{Mention the implementation and the successful outcome of the experiments}
% \direction{Potentially discuss options for future work}

In this work, we have formulated nonprehensile manipulation planning as a reinforcement learning problem. Concretely, we modeled the task with relevant rewards and trained a deep $Q$-network to generate actions based on the learned policy. Additionally, we proposed  \emph{replay buffer control} as well as potential field-based \emph{informed action sampling} for efficient training data collection to facilitate the network convergence.

We quantitatively evaluated the trained network by testing its success rate at different training stages, the results showed that the performance of the network was steadily improved, and that the network training was significantly affected by the ratio of success episodes in the reply buffer. After the network is converged, it achieved a success rate of $85\%$ implying that it has  learned how to handle the task. The average number of actions needed to complete a task has shown that the network was able to optimize its reward outputs to improve the action effectiveness. In comparison to a human subject, we can conclude that the network has achieved comparable performance to the human while it is more conservative in path planning for collision avoidance. Additionally, we have qualitatively shown that the network is reactive and adaptive to uncertainties, such as the sudden changes of objects and target positions, low-friction coefficients and distraction objects.

In future work, we plan to adapt the behaviors learned in simulation to a real environment, where we have no knowledge about physical properties of the objects and the lighting conditions. For this, we also need to enable the network to learn how to adapt the image input from a real camera to an image that can be used by the deep $Q$-network trained in simulation. Additionally, we would like to integrate more sensors, such as tactile and depth sensors, into our system to enable the cross-modal sensing ability for the system to better understand the task space, so as to more robustly handle the uncertainties in the real world.

% Future works：\\
% 1. adaption layer\\
% 2. combination with more sensors, because sometimes even human cannot see clearly\\
% 3. combination with traditional planning method, because the advantage of traditional method is what the learning method lack. e.g. at begining use planning method once, in the process use learning method based on planning result.

\section*{ACKNOWLEDGEMENT}
This work was supported by the HKUST SSTSP project RoMRO (FP802), HKUST IGN project IGN16EG09, HKUST PGS Fund of Office of Vice-President (Research \& Graduate Studies) and Knut and Alice Wallenberg Foundation \& Foundation for Strategic Research.


%%%%%%%%%%%%%%%%%%%%%%%%%%%%%%%%%%%%%%%%%%%%%%%%%%%%%%%%%%%%%%%%%%%%%%%%%%%%%%%%

{\small
\bibliographystyle{ieeetr}
\bibliography{ref}
}
\end{document}